\UseRawInputEncoding
\documentclass[lettersize,journal]{IEEEtran}
\usepackage{amsmath,amsfonts}
\usepackage{algorithmic}
\usepackage{algorithm}
\usepackage{array}
\usepackage{textcomp}
\usepackage{stfloats}
\usepackage{url}
\usepackage{verbatim}
\usepackage{graphicx}
\usepackage{cite}
\usepackage{color}
\usepackage{subcaption}
\usepackage{cleveref} 
\usepackage{amsthm,amsmath,amssymb}
\usepackage{multirow}
\usepackage{url}
\usepackage{amssymb}
\usepackage{pifont}

\usepackage{mathrsfs}
\newcommand{\ie}{\textit{i}.\textit{e}., }
\newcommand{\eg}{\textit{e}.\textit{g}., }
\newcommand{\vs}{\textit{vs.} }

\pdfoutput=1

\begin{document}

\title{Dual-branch Hybrid Learning Network for Unbiased Scene Graph Generation}


\author{Chaofan Zheng, Lianli Gao, Xinyu Lyu, Pengpeng Zeng, Abdulmotaleb El Saddik,~\IEEEmembership{Fellow,~IEEE},  Heng Tao Shen,~\IEEEmembership{Fellow,~IEEE}
\thanks{Chaofan Zheng, Lianli Gao, Xinyu Lyu, Pengpeng Zeng and Heng Tao Shen are with the
Future Media Center and School of Computer Science and Engineering, University of Electronic Science and Technology of China, Chengdu,
China, 611731. \\
Abdulmotaleb El Saddik is with Multimedia Communications Research Laboratory and School of Electrical Engineering and Computer Science, University of Ottawa, Ottawa, ON K1N 6N5, Canada. \\ 
Corresponding author: Lianli Gao. E-mail: lianli.gao@uestc.edu.cn}}

\maketitle

\begin{abstract}

The current studies of Scene Graph Generation (SGG) focus on solving the long-tailed problem for generating unbiased scene graphs.
However, most de-biasing methods over-emphasize the tail predicates and underestimate head ones throughout training, thereby wrecking the representation ability of head predicate features. Furthermore, these impaired features from head predicates harm the learning of tail predicates.
In fact, the inference of tail predicates heavily depends on the general patterns learned from head ones, \eg ``standing on'' depends on ``on''. Thus, these de-biasing SGG methods can neither achieve excellent performance on tail predicates nor satisfying behaviors on head ones.
To address this issue, we propose a Dual-branch Hybrid Learning network (DHL) to take care of both head predicates and tail ones for SGG, including a Coarse-grained Learning Branch (CLB) and a Fine-grained Learning Branch (FLB).
Specifically, the CLB is responsible for learning expertise and robust features of head predicates, while the FLB is expected to predict informative tail predicates.
Furthermore, DHL is equipped with a Branch Curriculum Schedule (BCS) to make the two branches work well together.
Experiments show that our approach achieves a new state-of-the-art performance on VG and GQA datasets and makes a trade-off between the performance of tail predicates and head ones. 
Moreover, extensive experiments on two downstream tasks (\ie Image Captioning and Sentence-to-Graph Retrieval) further verify the generalization and practicability of our method.
Our code is available at \url{https://github.com/aa200647963/SGG-DHL/}.

\end{abstract}
\begin{IEEEkeywords}
Scene Graph Generation, Vision and Language, Visual Understanding
\end{IEEEkeywords}

\section{Introduction}
\label{sec:intro}
\IEEEPARstart{T}{he} ability to transform unstructured data (\eg image) into structured data (\eg graph) is a fundamental human capability that children are taught from childhood. To formalize this unique ability, the task of Scene Graph Generation (SGG) is developed to simulate human-like interactions between vision and language. The ultimate target of this task is to produce a visually-grounded and relational-accurate graph, which covers most objects in an image and describes the relationships between them. Such structured graph can serve high-level visual-and-language tasks such as Visual Question Answering~\cite{vqa3,vqa2,vqa1}, Image Captioning~\cite{cap1,cap2}, and Image Retrieval~\cite{imgtrv1,imgtrv2}. 

\begin{figure}[t]
  \center
  \includegraphics[width=1\linewidth]{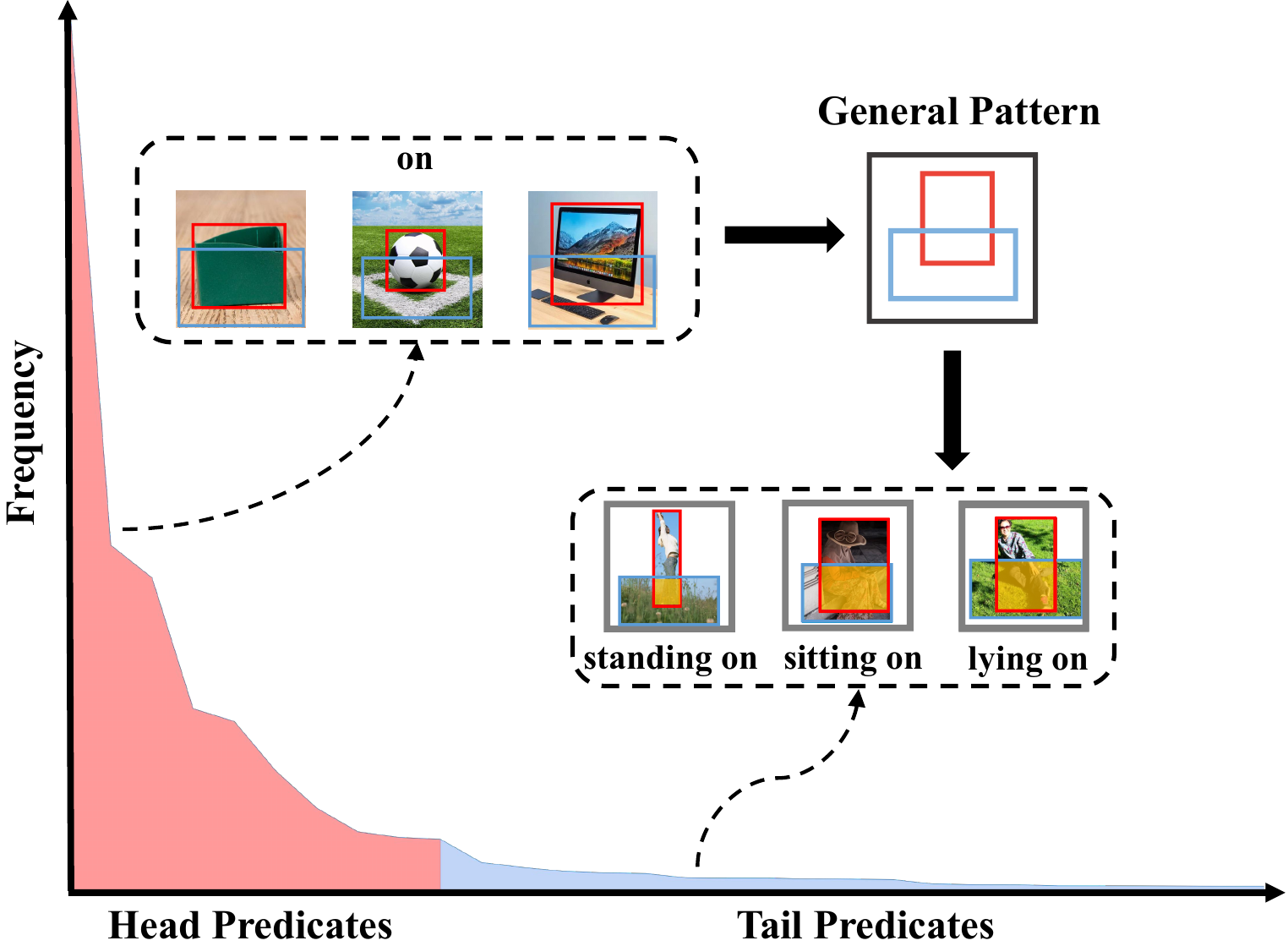}
\caption{The illustration of head predicate providing general pattern to tail predicate. The head predicate ``on" describes an object on top of another object, which shares the general pattern with tail predicates ``standing on'', ``sitting on'', and ``lying on''.
}
\label{fig:intro}
\end{figure}

The pipeline of existing general SGG models~\cite{sgg:graphrcnn,sgg:imp,sgg:motifs,sgg:vctree,sgg:kern} is: firstly extracting a set of object proposals and corresponding features, then aggregating the interaction information of these object proposals to construct contextual features for predicting the pairwise object relationships.
Although these methods have made great efforts to improve context representations, the generated scene graphs are far from satisfactory due to the long-tailed data distribution in Visual Genome~\cite{data:vg}, where only a few predicates have abundant samples, while most  have a few samples.
The predictions of SGG model trained with heavily imbalanced data distribution are biased towards the head predicates with coarse-grained descriptions.
Hence, it results in less informative scene graphs, which cannot provide rich semantic representations for downstream tasks. 

\begin{figure}[t]
    \centering
    \includegraphics[width=1\linewidth]{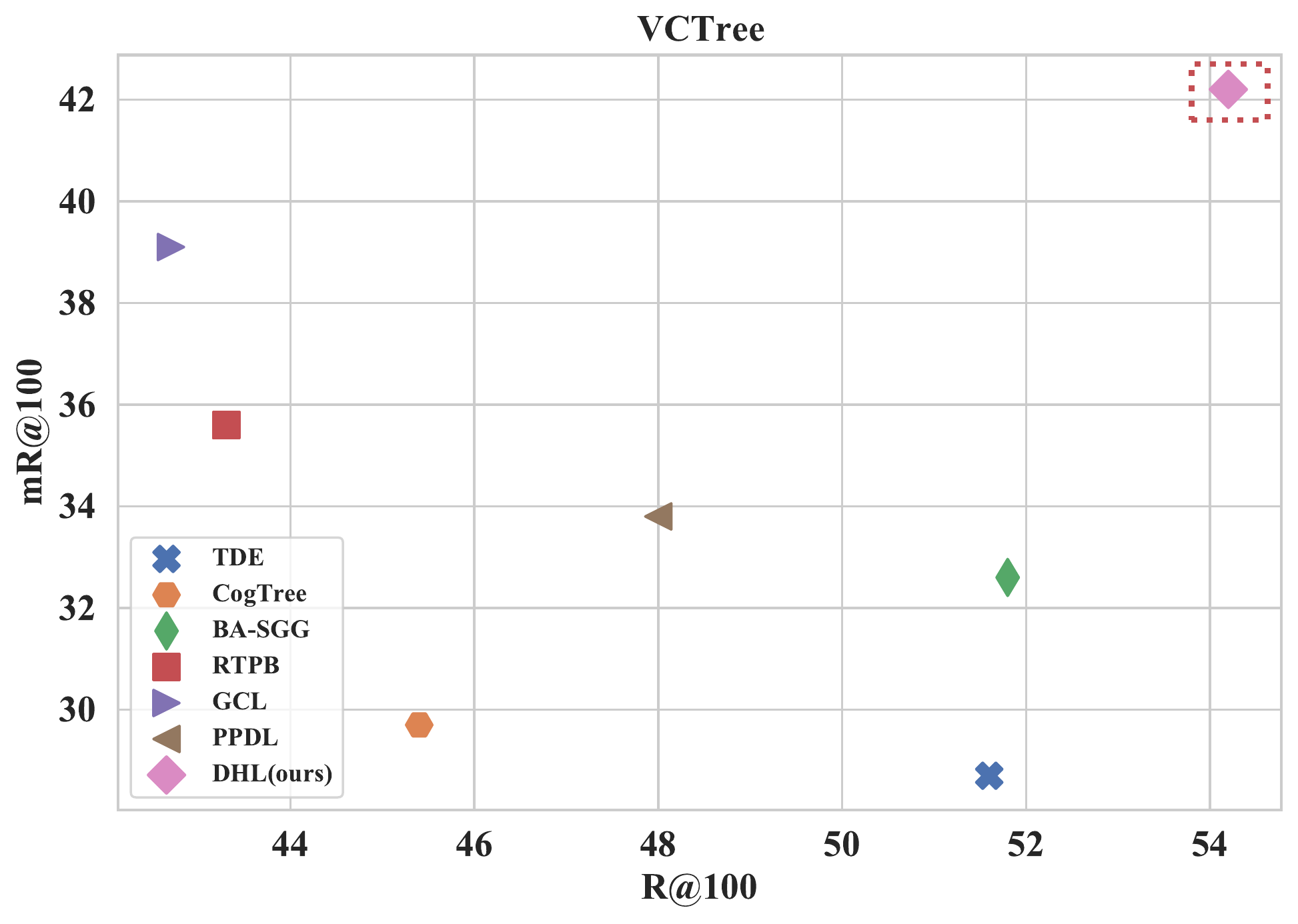}
    \caption{The illustration of performance distribution of different de-biasing methods for R@100 and mR@100 on the PredCls task. The baseline model is VCTree~\cite{sgg:vctree}.}
    \label{fig:performance_distribution_vctree}
\end{figure}

To mitigate the above issue, various de-biasing SGG methods~\cite{sgg:acbs,sgg:cogtree,sgg:fgpl,sgg:ba-sgg,sgg:gcl,sgg:tde,sgg:pcl} have been proposed. 
The mainstream methods can be roughly divided into two types: 1) re-sampling, which cuts down the head samples or repeats the tail samples to balance the distribution of training data, \eg GCL~\cite{sgg:gcl}, BA-SGG~\cite{sgg:ba-sgg}.
2) re-weighting: which assigns different weights to different predicates according to the predicate correlation, predicate frequency, or structured predicate prior to re-weight the contribution in the loss function, \eg CogTree~\cite{sgg:cogtree}, RTPB~\cite{sgg:rtpb}, PPDL~\cite{sgg:ppdl}.
Besides, TDE~\cite{sgg:tde} proposes a counterfactual causality method to distinguish the frequency biases in the training phase.
Although alleviating the imbalance problem to some extent, these de-biasing SGG methods still have adverse impacts.
For instance, in Fig.~\ref{fig:performance_distribution_vctree}, these methods fail to achieve an excellent performance of tail predicates and satisfactory behaviors of the head ones.
The potential reason is that they underestimate the head predicates throughout the whole training process, thereby wrecking the representation ability of the head predicates features and further affecting the learning of tail predicates.


Intuitively, head predicates share the general patterns with tail ones, which provides fundamental effects in recognizing tail predicates. As shown in Fig.~\ref{fig:intro}, the head predicate ``on'' may benefit the learning of the tail predicates, ``standing on'', ``sitting on'', and ``lying on''. 
Therefore, handling the biased problem in SGG requires the robust learning of head predicates.
In particular, the head predicates describe geometric (\eg “on”, “under”, “near”) or possessive (\eg “has”, “wearing”) relationships between objects, which are easy to learn in terms of spatial coordinates or co-occurrence patterns. However, the tail predicates are context-specific (\eg “eating”, “flying in”), which is more challenging for SGG models due to the need for conscious reasoning about contextual information. 
Thus, in conjunction with the above analysis, we can utilize a curriculum learning strategy, which first explores the general patterns from the head (easy) predicates and then gradually focuses on learning the tail (hard) predicates.

To this end, in this paper, we propose a \textbf{Dual-branch Hybrid Learning network (DHL)}, which takes care of both head predicates and tail predicates.
Specifically, our DHL consists of a \textbf{Coarse-grained Learning Branch (CLB)} and a \textbf{Fine-grained Learning Branch (FLB)}.
The CLB, optimized with the standard cross-entropy loss, is responsible for learning expertise and robust features of head predicates.
The FLB, optimized with the \textbf{Curriculum Re-weighting Mechanism (CRM)}, is designed to predict more informative tail predicates for generating practical scene graphs. 
After that, to make the most of the strengths of these two branches, we transfer the expertise in head predicates from CLB to FLB by knowledge distillation and design a \textbf{Branch Curriculum Schedule (BCS)} to control the model to learn the CLB first and then gradually pay attention to the FLB.
In specific, for the FLB, our \textbf{Curriculum Re-weighting Mechanism (CRM)} regulates it to learn the head (easy) predicates first and then gradually focus on the learning of tail (hard) ones by \textbf{Predicate Curriculum Schedule}.
Besides, we devise a \textbf{Semantic Context Module (SCM)} that fully explores and utilizes the context to correct unreliable predicate predictions of FLB to make the model more stable. 
Equipped with our DHL, VCTree~\cite{sgg:vctree} effectively balances the performance of head predicates and tail ones, and both outperform existing de-biasing methods, as shown in Fig~\ref{fig:performance_distribution_vctree}.

This paper reinforces the preliminary version of our work~\cite{sgg:icme} with a novel Dual-branch Hybrid Learning network architecture and insight analysis of each key component. 
The extensive contributions within this paper are four-folds. 
First, more comprehensive related works have been discussed in Sec.~\ref{sec:related_work}.
Second, we propose a Dual-branch Hybrid Learning network architecture to take care of both head predicates and tail predicates, which consists of a Coarse-grained Learning Branch that learns expertise and robust feature of head predicates and a Fine-grained Learning Branch that predicts more informative tail predicates.
In practice, we design a Branch Curriculum Schedule to take full advantage of these two branches. 
Third, compared with our previous method~\cite{sgg:icme}, DHL obtains significant performance improvements in the R@K metrics, such as the Transformer model, from 43.4 to 51.1, of up to 19\% in R@100 under the PredCls task. 
Finally, we conduct experiments on a more complex GQA~\cite{data:gqa} dataset to verify the generalization ability of our method and validate the practicability of generated scene graphs on Sentence-to-Graph Retrieval and Image Captioning tasks.

In summary, the main contributions of our work are three-folds:
\begin{itemize}
    \item
    We propose a novel Dual-branch Hybrid Learning network (DHL) to take care of both head predicates and tail ones, consisting of a Coarse-grained Learning Branch (CLB) for learning expertise and robust features of head predicates and a Fine-grained Learning Branch (FLB) for predicting informative tail predicates.
    Moreover, we devise a Branch Curriculum Schedule (BCS) to adjust the training process of the two branches. 
    
    \item
    We introduce a Curriculum Re-weighting Mechanism (CRM) to optimize the FLB, which regulates the FLB to learn the head predicates first and then progressively focus on the tail ones. Moreover, we devise a Semantic Context Module to correct out-of-context predictions of FLB to make the model more stable.
    
    \item
    Extensive experimental results demonstrate that our DHL dramatically boosts the performance of the baseline models on VG dataset and GQA dataset. For instance, our method improves Motifs by 163.9\%,
    VCTree by 142.5\%, and Transformer by 147.7\% in mR@100 under the PredCls task on VG dataset.  
    Furthermore, we conduct experiments on Sentence-to-Graph Retrieval and Image Captioning tasks, demonstrating the practicability of our DHL.
\end{itemize}

    
    

\section{RELATED WORK}
\label{sec:related_work}
\subsection{Scene Graph Generation}
Scene Graph Generation (SGG) targets producing a graphical summary of an image that benefits numerous downstream visual understanding tasks.
Early works~\cite{sgg:vrd,sgg:independent1} detect objects and relationships via independent networks, but they ignore the rich visual context information.  
To tackle this shortcoming, later, some works~\cite{sgg:gps,sgg:vctree,sgg:os-sgg,sgg:motifs,sgg:kern,sgg:imp,sgg:graphrcnn,sgg:runet,sgg:hlnet} devise various powerful context aggregation architectures to encode the context information, such as BiLSTMs~\cite{sgg:motifs}, graph neural networks~\cite{sgg:graphrcnn}, and dynamic tree structure~\cite{sgg:vctree}. 
Besides, some other works attempt to inject linguistic and human prior knowledge into the SGG model~\cite{sgg:vrd,sgg:motifs,sgg:kern,sgg:os-sgg,sgg:bridgeKnowledge,sgg:bridgeknowlegde2} for better improvements.
Although these methods greatly improved context representations, the generated scene graphs are far from satisfactory due to the biased data distribution.
To tackle the biased problem, some cleverly designed loss functions~\cite{sgg:pcpl,sgg:cogtree,sgg:rtpb,sgg:fgpl, sgg:pcl,sgg:ppdl,sgg:ebm,sgg:nice} and re-sampling strategies~\cite{sgg:ba-sgg,sgg:bgnn,sgg:acbs,sgg:gcl} have been introduced to generate unbiased scene graphs.~\cite{sgg:pcpl} proposes a flexible re-weighting method that utilizes the correlation among predicate classes to adaptively seek out appropriate loss weights.~\cite{sgg:cogtree} builds a hierarchical cognitive structure loss function from the cognition perspective to make the tail relationships receive more attention in a coarse-to-fine mode.~\cite{sgg:acbs} designs a novel alternating class-balanced sampling strategy to maintain the objects and predicates balanced. 
However, most of these de-biasing approaches emphasize the tail predicates and despise the head ones in the training process, damaging the learning of head predicates while further compromising the learning of tail predicates. 
In this work, we propose a Dual-branch Hybrid Learning network, which takes care of both head predicates learning and tail predicates learning, balancing the performance of head predicates and tail predicates well. 

\subsection{Long-Tailed Recognition}
Recently, the long-tailed-based recognition problem has attracted significant attention. 
A classic method to deal with the long-tailed problem is re-sampling, which can achieve a balanced sample space by adjusting the sampling ratio of different classes. It is divided into two types: under-sampling~\cite{undersamp1,undersamp2} and over-sampling~\cite{oversamp1,undersamp2}. 
The under-sampling strategy discards the sufficient head samples, wasting data and impairing the model's generalization ability.
The over-sampling strategy repeats the infrequent tail samples, leading to over-fitting of the tail classes. 
Another effective method is to re-weight~\cite{loss1,loss:cbloss,loss:focalloss} the loss function, which usually assigns large weights for the tail classes in the loss function. 
A simple re-weighting way is to set the weight of a class to the inverse of its proportional, but this method leads to weak performance on head classes. Subsequently, ~\cite{loss:cbloss} proposes to adopt the effective number of samples instead of frequency to alleviate this problem. 
These methods assume that the categories are independent, ignoring the intrinsic associations between categories.
Unlike these methods, our approach fully mines and utilizes the general patterns from head predicates to learn the tail predicates better.

\section{METHOD}

\begin{figure}[t]
  \center
  \includegraphics[width=1\linewidth]{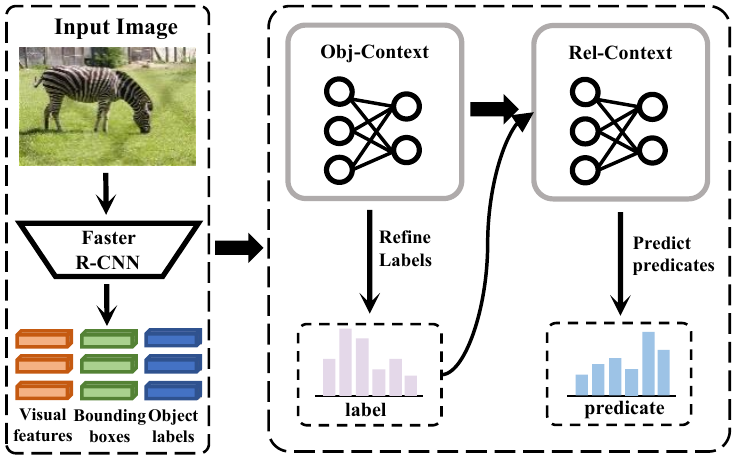}
\caption{The illustration of the pipeline of vanilla Scene Graph Generation Methods.
}
\label{fig:s-framework}
\end{figure}

\begin{figure*}[t]
 \centering
 \includegraphics[width=1.0\linewidth]{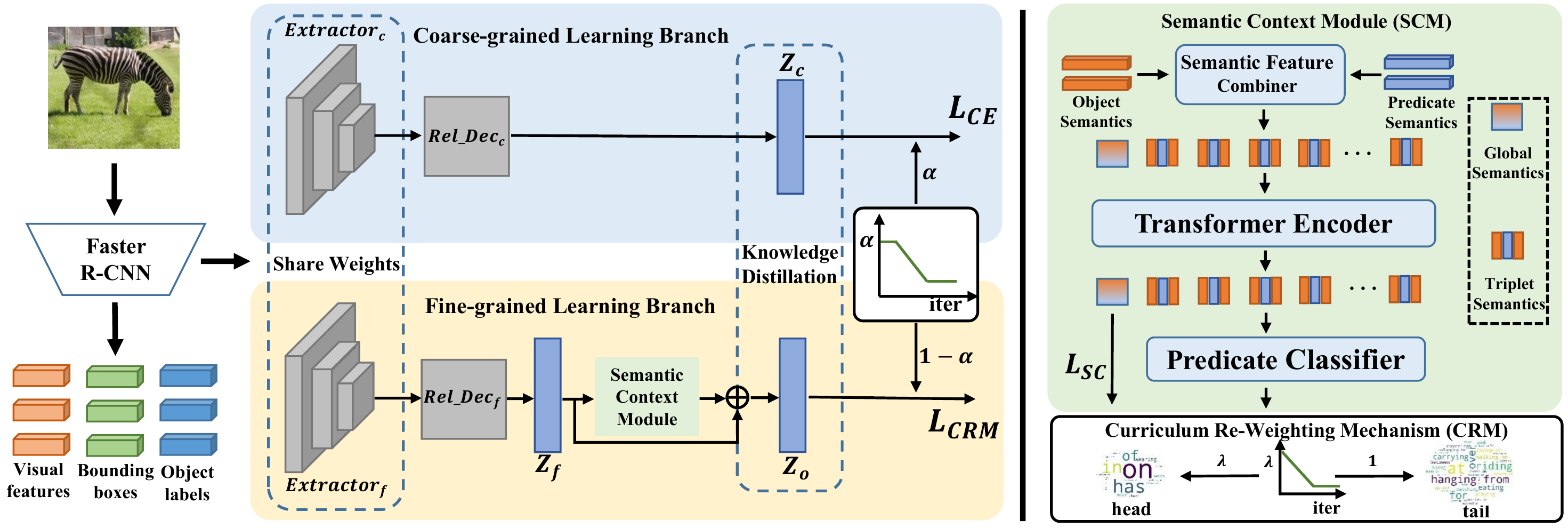} 
 \caption{Overview of the Dual-branch Hybrid Learning network. 
 The Faster R-CNN is used to obtain visual features, bounding boxes, and object labels of object proposals.
 Then, these features are fed into a Coarse-grained Learning Branch optimized with cross-entropy loss and a Fine-grained Learning Branch optimized with Curriculum Re-weighting Mechanism.
 Moreover, knowledge distillation is used to impart the expertise of CLB in head predicates to FLB, and a Branch Curriculum Schedule is used to adjust the learning weights between the two branches.
 In particular, the Curriculum Re-weighting Mechanism gradually transfers the learning focus of FLB from head predicates to tail predicates by Predicate Curriculum Schedule. 
 Besides, we use the Semantic Context Module to correct out-of-context predictions and ensure semantic consistency.
 }
 \label{fig:kjt}
\end{figure*}

\subsection{Overview}
\label{sec:DBHLM}
As shown in Fig.~\ref{fig:kjt}, our Dual-branch Hybrid Learning network (DHL) consists of a Coarse-grained Learning Branch (CLB, Sec.~\ref{sec:clb}) and a Fine-grained Learning Branch (FLB, Sec.~\ref{sec:flb}).
The Coarse-grained Learning Branch, optimized with standard cross-entropy loss function, is responsible for learning expertise and robust features of head predicates.
The Fine-grained Learning Branch is optimized with our Curriculum Re-weighting Mechanism (CRM, Sec.~\ref{sec:CRM}), which prefers to predict more informative tail predicates. 
However, this weakens its ability to recognize head predicates. 
Thus, we impart the expertise of CLB in head predicates to FLB by knowledge distillation (Sec.~\ref{sec:kd}).
Besides, we design a Semantic Context Module (SCM, Sec.~\ref{sec:SCM}) to correct the unreliable predicate predictions in FLB.
Finally, we use a Branch Curriculum Schedule (BCS, Sec.~\ref{sec:bcs}) to adjust the training focus between the two branches to make them work well together.
During inference, we only need to use the FLB for the unbiased scene graph generation.

\subsection{Coarse-grained Learning Branch}
\label{sec:clb}
The Coarse-grained Learning Branch is built on the vanilla SGG models, \eg Motifs~\cite{sgg:motifs}, VCTree~\cite{sgg:vctree}, and Transformer~\cite{transformer,sgg:sggbenchmark}.
The common pipeline of these vanilla SGG models is shown in Fig~\ref{fig:s-framework}.

Firstly, the Faster R-CNN~\cite{fasterrcnn} framework is used to obtain object proposals and corresponding features, including:

\begin{itemize}
    \item A set of bounding boxes $ B=\{b_1, b_2, ... , b_n\} $, where $b_i \in \mathbb{R}^4$  denotes the spatial locations of detected regions.
    \item A set of object proposals' visual features $ V = \{v_1, v_2, ..., v_n\} $, where $v_i \in \mathbb{R}^{4096}$.  
    \item A set of probability distributions of object labels  $ L = \{l_1, l_2, ... , l_n\}$, where $l_i \in \mathbb{R}^{N_O+1}$, $N_O$ is the number of object classes. 
    \item A set of features of union box of a pair of proposals $i$ and $j$, $U = \{u_{i,j} | i,j \in n, i \neq j \}$, where $u_{i,j} \in  \mathbb{R}^{4096}$.
\end{itemize}

Then, these features are fed into a context message passing module consisting of the Object Context Module and Relation Context Module. The Object Context Module includes Object Encoder and Object Decoder, and the Relation Context Module includes Relation Encoder and Relation Decoder. 
A brief introduction of context message passing module is given below:
\begin{enumerate}
    \item[(1)]
    \textbf{Object Encoder}. It is used to construct contextual representations of objects for further label prediction, which is formed as:
    \begin{equation}
        C = Obj\_Enc([v_i;W_b b_i;Emb(l_i)]_{i=1,...,n}),
    \end{equation}
    where $C=\{c_1, c_2,..., c_n\}$ is a set of contextual object representations, $W_b$ is a trainable parameter matrix that maps the bounding box coordinates to $\mathbb{R}^{100}$, $Emb(\cdot)$ is a pre-trained word embedding model (GloVe) to convert $l_i$ to its semantics representation, and $[\cdot;\cdot]$ denotes the concatenation operation.
    \item[(2)]
    \textbf{Object Decoder}. It is used to obtain the refined object labels based on the contextual object features $C$, which is calculated as:
    \begin{equation}
        L' = Obj\_Dec(C),
    \label{eq:lp}
    \end{equation}
    where $L'=\{l'_1, l'_2,...,l'_n\}$ is the refined probability distributions of object labels. 
    \item[(3)]
    \textbf{Relation Encoder}. To further enhance the contextual information of object representations and use the refined object labels, the Relationship Encoder is proposed, which is calculated as:
    \begin{equation}
        C' = Rel\_Enc([c_i;Emb(l'_i)]_{i=1,...,n}),
    \end{equation}
    where $C'=\{c'_1,c'_2,...,c'_n\}$ is the final context representation of objects for predicate prediction.
    \item[(4)]
    \textbf{Relation Decoder}. It is used to predict the relation predicate between two objects, which is calculated as:
    \begin{equation}
        z_{i,j} = Rel\_Dec([c'_i;c'_j;u_{i,j}]) + W_{o_i,o_j},
    \end{equation}
    where $W_{o_i,o_j}$ is a prior bias vector specific to the subject and object, $u_{i,j}$ is the union feature of the objects $i$ and $j$ ,$o_i$ and $o_j$ are the category indices of objects $i$ and $j$, and $z_{i,j}$ is the predicted predicate logits between object $i$ and $j$.
\end{enumerate}
Furthermore, we give a sample formulation of the context message passing module in the following:
\begin{equation}
\begin{aligned}
    C'_c & = Extractor_c(B,V,L) \\
    Z_c & = Rel\_Dec_c(C'_c,U,W_{bias})
\end{aligned},
\end{equation}
where $Extractor_c(\cdot,\cdot,\cdot)$ is a relation feature extractor that consists of the object context module and relation encoder, $C'_c$ is a set of final contextual features of objects, $Rel\_Dec_c(\cdot,\cdot,\cdot)$ is a relation decoder,
$W_{bias}$ is the statistic prior bias of the training set, and $Z_c$ is a set of predicates logits.

Finally, the standard cross-entropy loss function is used to optimize above method. Given the predicted predicate logits $z= [z_1,z_2,...,z_{N_R+1}]$ ($N_R$ predicate classes and a background class) and ground truth label $y=[y_1,y_2,...,y_{N_R+1}]$, in which $y_i$ equals 1 or 0, the cross-entropy loss function is formed as:

\begin{equation}
\label{eq:loss_ce}
  \mathcal{L}_{CE}(z,y) =  -  \sum_{i=1}^{N_R+1} y_i \log{\frac{\exp(z_i)}{\sum_{j=1}^{N_R+1} \exp(z_j)}}.
\end{equation}
Due to the imbalanced data distribution that head predicates have abundant samples, the training process of CLB is dominated by the head predicates.
Thus, it can obtain robust features and expertise in head predicates.

\subsection{Fine-grained Learning Branch}
\label{sec:flb}
The Fine-grained Learning Branch is also constructed on the basis of the vanilla SGG model, which is optimized with the Curriculum Re-weighting Mechanism. It contains the following operations:
\begin{equation}
\begin{aligned}
    C'_f & = Extractor_f(B,V,L) \\
    Z_f & = Rel\_Dec_f(C'_f,U,W_{bias}) \\ 
    \widetilde{Z} & = SCM(Z_f, L') \\
    Z_o & = Z_f + \widetilde{Z}
\end{aligned},
\label{eq:flb}
\end{equation}
where $L'$ is the refined probability distributions of object labels mentioned in Eq. (\ref{eq:lp}), $SCM(\cdot, \cdot)$ is our Semantic Context Module,
$\widetilde{Z}$ is the modified predicate logits set used to correct the unreliable predictions in $Z_f$, 
and $Z_o$ is the final predicate logits set for output. 
It should be noted that the relation feature extractors $Extractor_f(\cdot,\cdot,\cdot)$ and $Extractor_c(\cdot,\cdot,\cdot)$ share parameters.
Thus, FLB shares the robust features of head predicates learned by the CLB.
Next, we describe Curriculum Re-weighting Mechanism and Semantic Context Module in detail.

\subsubsection{Curriculum Re-weighting Mechanism}
\label{sec:CRM}
As previously stated, head predicates share the general patterns with tail ones, which provides fundamental effects in recognizing tail predicates.
Therefore, we devise a Curriculum Re-weighting Mechanism to fully explores the general patterns from head predicates to learn the tail predicates better. 
It regulates the model to learn the head predicates first and then gradually focus on the tail predicates by Predicate Curriculum Schedule.
The revised loss function is as follows:
\begin{equation}\label{Eq:3.2}
  \mathcal{L}_{CRW}(z,y) =  - \sum_{i=1}^{R+1}\lambda_i w_i y_i \log{\frac{\exp(z_i)}{\sum_{j=1}^{R+1} \exp(z_j)}},
\end{equation}
where $w_i$ is the weight of class $i$ computed with a re-weighting method~\cite{loss:cbloss} . The trade-off factor \textbf{$\lambda_i$} of Predicate Curriculum Schedule is defined as:
\begin{equation}
\lambda_i=
\begin{cases}
max(\varphi_1(k), \beta_1)& \text{if $i \in H$}\\
1& \text{otherwise}
\end{cases},
\end{equation}
where $H$ is the set of head predicate indexes, and those with more than $M$ samples are considered as head predicates. 
$\varphi_1(k)$ is a \textbf{Schedule Function} decreasing from 1 to 0, representing the learning ``attention'' weights allocated to the head predicates. 
In order to prevent the forgetting of head predicates, a threshold hyper-parameter $\beta_1 \in [0,1]$ is used to avoid zero weights for them.
The $\varphi_1(k)$ is defined as:
\begin{equation}
\label{eql-4}
  \varphi_1(k) = 1 - \frac{k - K_1}{K - K_1},
\end{equation}
where $k$ is the current training iteration, $K$ refers to the total training iterations, and $K_1$ is a intermediate training iteration introduced in Eq.~(\ref{eq:alpha}).

\subsubsection{Semantic Context Module}
\label{sec:SCM}
For the scene graph of an image, the choice of a relation triple is highly dependent on the context provided by the other triples. 
However, most existing SGG methods~\cite{sgg:motifs,sgg:vctree,sgg:ba-sgg,sgg:tde,sgg:graphrcnn,sgg:cogtree} are independent and parallel to predict all relation predicates in an image, which ignores the correlation between the relation triples. This leads to a risk that the predicted relation predicate may deviate from the context.
To alleviate this problem, we design a Semantic Context Module, which takes the local semantic representations of the relation triplets and the global semantic representation of the whole graph as inputs and generates the contextual semantic representations. 
The local contextual representations can correct out-of-context predictions in each triplet individually.
The global contextual representation measures the semantic gap between the generated scene graph and the ground truth.
It is described in more detail below.

We process the predicate logits set $Z_f$ to obtain a set of predicate probability distributions $\{p_1, p_2,...,p_N\}$, where $N$ is the total number of relations in an image and $p_i \in \mathbb{R}^{N_R+1}$.
Next, we map each probability distribution of predicates and objects to a 200-dimensional vector with a pre-trained word embedding model (GloVe) to obtain the predicate semantic representation $s^p$ and object semantic representation $s^o$.
Then, the predicate semantic representation is concatenated with the corresponding semantic representations of subject and object to get the relation triplet semantic representation, as follows:
\begin{equation}
    s_i^r = [s_i^{o,s}; s_i^p; s_i^{o,o}]W, W \in \mathbb{R}^{600 \times D},
\label{eq:sc}
\end{equation}
where $s_i^{o,s}$ and $s_i^{o,o}$ are the subject and object semantic representations corresponding to the $s_i^p$, and $W$ is a trainable linear projection that maps the concatenated semantic representation to $D$ dimensions. In addition, we add an extra global representation $s_{global}$ as the semantic representation of the whole scene graph, which is defined as follows:
\begin{equation}
    s_{global} = \frac{1}{N} \sum_{i=1}^{N}{s_i^r}.
\end{equation}
The same processing is performed on the ground truth relation triplets to get $t_i^r$ and $t_{global}$. 
Based on these preparations, a vanilla Transformer~\cite{transformer} Encoder is used to construct contextual semantic representations. For simplicity, we denote the vanilla Transformer Encoder as \textbf{$Trans\_Enc(\cdot)$}.
The input of $Trans\_Enc(\cdot)$ is  $S^r=\{s_1^r,s_2^r,...,s_N^r,s_{global}\}$ and the contextual semantic representation, $\widetilde{S^r}$, is computed as follows:
\begin{equation}
    \widetilde{S^r} = Trans\_Enc(S^r),
\end{equation}
where $\widetilde{S^r}=\{\widetilde{s^r_1},\widetilde{s_2^r},...,\widetilde{s_N^r},\widetilde{s}_{global}\}$. 
The ground truth contextual semantic representation $\widetilde{T^r}$ is also obtained with  $Trans\_Enc(\cdot)$. Then,  $\widetilde{s}_{global}$ and $\widetilde{t}_{global}$ are used to compute the semantic gap between the generated scene graph and the ground truth, and a mean-squared loss is used to minimize it:
\begin{equation}
    \mathcal{L}_{SC} = \frac{1}{D} {\Vert\widetilde{s}_{global} - \widetilde{t}_{global}\Vert}^2,
\label{eq:sc_loss}
\end{equation}
where $D$ is the same as in Eq.~(\ref{eq:sc}).  
Besides, these triples semantic representations $\{\widetilde{s^r_1},\widetilde{s_2^r},...,\widetilde{s_N^r}\}$ are used for predicate classification to obtain modified predicate logits set $\widetilde{Z}$. 
Afterward, we add modified predicate logits $\widetilde{Z}$ to original predicate logits $Z_f$ to correct the out-of-context predictions. The final predicate logits $Z_o$ is computed as follows:
\begin{equation}
    Z_o = Z_f  + \widetilde{Z}.
\label{eq:zr}
\end{equation}



\subsection{Knowledge Distillation between CLB and FLB}
\label{sec:kd}
The FLB excels at predicting tail predicates, which inevitably weakens its ability to identify head predicates. 
However, the CLB is experienced in head predicates, so we can distill the expertise of it to FLB.
This improves the reliability of head predicates classification of FLB directly. The formula of knowledge distillation is as follows:
\begin{equation}
    \mathcal{L}_{KD}(p^T,q^T) = - \sum_{i}^{N_R+1} p_i^T \log(q_i^T),
    \label{eq:loss_kd}
\end{equation}
where $p^T$ and $q^T$ are the soft probability distributions about the predicate logits $z^c$ and $z^o$. They are computed as:
\begin{equation}
\begin{aligned}
p_i^T  &= \frac{\exp(z^c_i/\tau)}{\sum_{j}^{N_R+1}\exp(z^c_j/\tau)} \\ 
q_i^T  &= \frac{\exp(z^o_i/\tau)}{\sum_{j}^{N_R+1}\exp(z^o_j/\tau)}
\end{aligned},
\label{eq:kd_t}
\end{equation}
where $\tau$ is a temperature hyper-parameter. It is worth noting that we only perform the knowledge distillation between the head predicates. 

\subsection{Branch Curriculum Schedule}
\label{sec:bcs}
In order to reasonably plan the training of these two branches, we design a Branch Curriculum Schedule to adjust the learning weights between CLB and FLB by a trade-off factor $\alpha$. $\alpha$ is defined as:
\begin{equation}
\label{eq:alpha}
    \alpha = 
    \begin{cases}
    1 & \text{if $k \leq K_1$} \\ 
    \max(\varphi_2(k),\beta_2) & \text{if $K_1 < k \leq K_2$} \\ 
    \beta_2 & \text{if $k > K_2$}
    \end{cases}, 
\end{equation}
where $k$ is the current training iteration, $K_1$ and $K_2$ are the intermediate training iterations.
$\beta_2 \in [0,1]$ is a threshold hyper-parameter used to avoid assigning zero weight to CLB. $\varphi_2(k)$ is another \textbf{Schedule Function} decreasing from 1 to 0 with the input iteration $k$, which is defined as:
\begin{equation}
    \varphi_2(k) = 1 - \frac{k-K_1}{K_2-K_1}.
\end{equation}
The joint loss function for these two branches is:
\begin{equation}
    \mathcal{L}_{hybird} = \alpha \cdot \mathcal{L}_{CE} + (1 - \alpha) \cdot \mathcal{L}_{CRM}.
\end{equation}
In the training process, the loss $\mathcal{L}_{CE}$ of CLB is multiplied by $\alpha$, and the loss $\mathcal{L}_{CRM}$ of FLB is multiplied by $1-\alpha$. 
According to the definition of $\alpha$, the training process is divided into three phases. 
At the first training phase ($k \leq K_1$), $\alpha$ is equal to 1, so only CLB is trained to learn expertise and robust features of coarse-grained head predicates.
At the second phase ($K_1 < k \leq K_2$), $\alpha$ gradually decreases with the training. 
The model's learning focus shifts from CLB to FLB. 
Therefore, from this stage, FLB can share the robust features and expertise of head predicates learned by CLB.
At the third phase ($k > K_2$), $\alpha$ equals $\beta_2$. In this stage, CLB and FLB have constant learning weights $\beta_2$ and $1-\beta_2$ and train simultaneously until the end.
With the threshold of $\beta_2$, DHL avoids damaging the robust learned features when focusing on the tail predicates at later training periods.
Finally, the total loss for our DHL is computed as follows: 
\begin{equation}
    \mathcal{L}_{total} = \mathcal{L}_{hybrid} + \mathcal{L}_{SC} + \mu \cdot \mathcal{L}_{KD},
\end{equation}
where $\mu$ is a balancing factory.

\section{EXPERIMENTS}
In this section, we first introduce the experimental settings of scene graph generation and two downstream tasks (sentence-to-graph retrieval and image captioning), including datasets, evaluations, and implementation details. 
Then, we perform extensive experiments to demonstrate the superiority of our method and its effectiveness in downstream tasks. 
Finally, we conduct ablation studies and hyper-parameter analyses on the proposed components and present visualization results.
\subsection{Datasets}
\noindent
\textbf{Scene Graph Generation.} 
We conduct the experiments on two datasets: Visual Genome (VG)~\cite{data:vg} and GQA~\cite{data:gqa}.
VG dataset is the widely used benchmark, composed of 108K images with average annotations of 38 objects and 22 relationships per image. In this paper, we follow previous works~\cite{sgg:tde,sgg:motifs,sgg:imp} and adopt the most popular split from~\cite{sgg:imp}, which contains the most frequent 150 object categories and 50 predicate categories. 
Moreover, the VG dataset is divided into a training set with 70\% of the images and a testing set with the remaining 30\%, and 5K images from the training set for validation. 
GQA dataset is constructed from the Visual Genome dataset but filters out inaccurate predicates and augments object and relation annotations. It contains 1,704 object categories and 311 predicate categories. Since some low-frequency predicates and objects are not filtered, it has a severer imbalance problem than VG dataset. We use the official split, which has about 75K images in the training set and 10K in the testing set. We further sample 5K images from the training set as the validation set.  

\noindent
\textbf{Sentence-to-Graph Retrieval.} 
For the Sentence-to-Graph Retrieval task, we follow the setting of~\cite{sgg:tde}, which contains 41K overlapping images between Visual Genome and MS-COCO~\cite{data:mscoco} Caption datasets. Moreover, these images are divided into a training set with 35K images and two testing sets with 1K/5K images named test-1K/test-5K. 

\noindent
\textbf{Image Captioning.} 
We use the most popular benchmark dataset MS-COCO~\cite{data:mscoco} for the image captioning task, which contains 123K images, each annotated with 5 different captions. We adopt the split provided by~\cite{cocosplit}, which contains 113K images in the training set, and 5K images, respectively, for testing and validation.

\subsection{Evaluations}
\noindent
\textbf{Scene Graph Generation.} 
We follow previous works~\cite{sgg:ba-sgg,sgg:tde,sgg:imp} to evaluate our method on three subtasks: 
(1) Predicate Classification (\textbf{PredCls}): given the ground-truth bounding boxes and object labels in an image, predict the relation predicate labels. 
(2) Scene Graph Classification (\textbf{SGCls}): given the ground-truth bounding boxes in an image, predict the object labels and the relation predicate labels. 
(3) Scene Graph Detection (\textbf{SGDet}): given an image, predict the scene graph from scratch. 
Moreover, in this work, we use three evaluation metrics: Recall@K (R@K), mean Recall@K (mR@K), and Mean@K (M@K). 
The R@K measures the proportion of correctly predicted relations in the top K relation predictions.
Due to the heavily biased training data, R@K is easily dominated by a few head predicates with abundant samples.
The mR@K computes the R@K for each predicate category and averages them.
Following previous works~\cite{sgg:tde,sgg:kern,sgg:ba-sgg,sgg:fgpl}, we use the mR@K as our primary evaluation metric.
The M@K is calculated by averaging the R@K and mR@K, reflecting the model's comprehensive performance.
In addition, we also use M@K as the basis for model selection in ablation studies.

\noindent
\textbf{Sentence-to-Graph Retrieval.} 
The R@K and mR@K reflect the performance of SGG models at the triplet level, while the Sentence-to-Graph Retrieval (S2GR) task reflects the evaluation of the generated scene graph at the graph level. 
For the S2GR task, the visual features of the image are abandoned, and the scene graph detected by SGDet is deemed the only information of one image. Then, the caption sentences of this image are used as queries to retrieve it.
In the testing stage, the Recall@20/50/100 are reported on the gallery size of 1K and 5K. 

\noindent
\textbf{Image Captioning.}
In order to further verify the practicability of scene graph as additional information for the downstream task, we construct an image captioning experiment, in which scene graphs are used as auxiliary information and combined with visual feature to generate captions.
Following the standard evaluation protocol, we use BLEU~\cite{bleu}, METEOR~\cite{meteor}, CIDER~\cite{cider} and SPICE~\cite{spice} as metrics to evaluate the quality of image captioning. 

\begin{table*}[htp] \centering 
\caption{Performance comparison between our method (DHL) and existing model-agnostic de-biasing methods on PredCls, SGCls and SGDet tasks of VG dataset with respect to mR@20/50/100 . The results of other methods are reported from the corresponding papers. Our re-implemented SGG models are denoted by the superscript $*$.}
\resizebox{\textwidth}{!}{
\begin{tabular}{cc|ccc|ccc|ccc}
\hline
\multirow{2}{*}{Model} &
\multirow{2}{*}{Method} & \multicolumn{3}{c|}{PredCls} & \multicolumn{3}{c|}{SGCls} & \multicolumn{3}{c}{SGDet} \\
\cline{3-11}
& & mR@20 & mR@50 & mR@100 & mR@20 & mR@50 & mR@100 & mR@20 & mR@50 & mR@100 \\
\hline
\multirow{9}{*}{Motifs} 
& baseline* & 11.5 & 14.6 & 15.8 & 6.5 & 8.0 & 8.5 & 4.8 & 6.2 & 7.1 \\
&TDE~\cite{sgg:tde}$\ _{\textit{CVPR'20}}$ & 18.5 & 25.5 & 29.1 & 9.8 & 13.1 & 14.9 & 5.8 & 8.2 & 9.8\\
&EBM~\cite{sgg:ebm}$\ _{\textit{CVPR'21}}$ & 14.2 & 18.0 & 19.5 & 8.2 & 10.2 & 11.0 & 5.7 & 7.7 & 9.3 \\
&CogTree~\cite{sgg:cogtree}$\ _{\textit{IJCAI'21}}$ & 20.9 & 26.4 & 29.0 & 12.1 & 14.9 & 16.1 & 7.9 & 10.4 & 11.8 \\ 
&BA-SGG~\cite{sgg:ba-sgg}$\ _{\textit{ICCV'21}}$ & 24.8 & 29.7 & 31.7 & 14.0 & 16.5 & 17.5 & 10.7 & 13.5 & 15.6 \\ 
&RTPB~\cite{sgg:rtpb}$\ _{\textit{AAAI'22}}$ & 28.8 & 35.3 & 37.7 & 16.3 & 19.4 & 20.6 & 9.7 & 13.1 & 15.5 \\ 
&FGPL~\cite{sgg:fgpl}$\ _{\textit{CVPR'22}}$& 24.3 & 33.0 & 37.5 & 17.1 & \underline{21.3} & \underline{22.5} & 11.1 & 15.4 & 18.2 \\
&GCL~\cite{sgg:gcl}$\ _{\textit{CVPR'22}}$ & \underline{30.5} & \underline{36.1} & \underline{38.2} & \underline{18.0} & 20.8 & 21.8 & \underline{12.9} & \underline{16.8} & \underline{19.3} \\
&PPDL~\cite{sgg:ppdl}$\ _{\textit{CVPR'22}}$ & - & 32.2 & 33.3 & - & 17.5 & 18.2 & - & 11.4 & 13.5 \\
& \textbf{DHL (ours)} & \textbf{32.9} & \textbf{39.1} & \textbf{41.7} & \textbf{19.8} & \textbf{23.1} & \textbf{24.1} & \textbf{13.9} & \textbf{17.8} & \textbf{20.7} \\ 
\hline
\multirow{9}{*}{VCTree} 
& baseline* & 12.1 & 15.3 & 17.4 & 8.2 & 10.3 & 11.5 & 5.2 & 6.7 & 7.9 \\ 
&TDE~\cite{sgg:tde}$\ _{\textit{CVPR'20}}$ & 18.4 & 25.4 & 28.7 & 8.9 & 12.2 & 14.0 & 6.9 & 9.3 & 11.1 \\ 
&EBM~\cite{sgg:ebm}$\ _{\textit{CVPR'21}}$ & 14.2 & 18.2 & 19.7 & 10.4 & 12.5 & 13.4 & 5.7 & 7.7 & 9.1 \\
&CogTree~\cite{sgg:cogtree}$\ _{\textit{IJCAI'21}}$ & 22.0 & 27.6 & 29.7 & 15.4 & 18.8 & 19.9 & 7.8 & 10.4 & 12.1 \\ 
&BA-SGG~\cite{sgg:ba-sgg}$\ _{\textit{ICCV'21}}$ & 26.2 & 30.6 & 32.6 & 17.2 & 20.1 & 21.2 & 10.6 & 13.5 & 15.7 \\ 
&RTPB~\cite{sgg:rtpb}$\ _{\textit{AAAI'22}}$ & 27.3 & 33.4 & 35.6 & 20.6 & 24.5 & 25.8 & 9.6 & 12.8 & 15.1 \\ 
&FGPL~\cite{sgg:fgpl}$\ _{\textit{CVPR'22}}$ & 30.8 & \underline{37.5} & \underline{40.2} & \underline{21.9} & \underline{26.2} & \underline{27.6} & \underline{11.9} & \underline{16.2} & \underline{19.1} \\
&GCL~\cite{sgg:gcl}$\ _{\textit{CVPR'22}}$ & \underline{31.4} & 37.1 & 39.1 & 19.5 & 22.5 & 23.5 & \underline{11.9} & 15.2 & 17.5 \\ 
&PPDL~\cite{sgg:ppdl}$\ _{\textit{CVPR'22}}$ & - & 33.3 & 33.8 & - & 21.8 & 22.4 & - & 11.3 & 13.3\\ 
& \textbf{DHL (ours)} & \textbf{33.3} & \textbf{40.0} & \textbf{42.2} & \textbf{23.2} & \textbf{26.9} & \textbf{28.2} & \textbf{13.5} & \textbf{17.4} & \textbf{20.0} \\ 
\hline
\multirow{4}{*}{Transformer} 
& baseline* & 12.6 & 15.9 & 17.2 & 7.7 & 9.8 & 10.5 & 5.6 & 7.9 & 9.0\\
& BA-SGG~\cite{sgg:ba-sgg}$\ _{\textit{ICCV'21}}$ & 26.7 & 31.9 & 34.2 & 15.7 & 18.5 & 19.4 & 11.4 & 14.8 & 17.1 \\ 
& FGPL~\cite{sgg:fgpl}$\ _{\textit{CVPR'22}}$ & \underline{27.5} & \underline{36.4} & \underline{40.3} & \underline{16.5} & \underline{21.6} & \underline{23.8} & \underline{13.2} & \underline{17.4} & \underline{20.3} \\
& \textbf{DHL (our)} & \textbf{34.5} & \textbf{40.4} & \textbf{42.6} & \textbf{20.4} & \textbf{24.2} & \textbf{25.3} & \textbf{13.9} & \textbf{18.2} & \textbf{21.0} \\
\hline
\end{tabular}
}
\label{tab:mR@k} 
\end{table*}

\begin{table*}[htp] \centering 
\caption{Comprehensive performance comparison of different de-biasing methods on VG dataset. The R@50/100, mR@50/100, and M@50/100 on PredCls, SGCls and SGDet tasks are reported. }
\resizebox{\textwidth}{!}{
\begin{tabular}{c|ccc|ccc|ccc}
\hline
\multirow{2}{*}{Model+Method} & \multicolumn{3}{c|}{PredCls} & \multicolumn{3}{c|}{SGCls} & \multicolumn{3}{c}{SGDet} \\
\cline{2-10}
& R@50 / 100 & mR@50 / 100 & M@50 / 100 &  R@50 / 100 & mR@50 / 100 & M@50 / 100 &  R@50 / 100 & mR@50 / 100 & M@50 / 100  \\
\hline
PCPL~\cite{sgg:pcpl}$\ _{\textit{ACM MM'20}}$  & 50.8 / 52.6 & 35.2 / 37.8 & 43.0 / 45.2 & 27.6 / 28.4 & 18.6 / 19.6 & 23.1 / 24.0 & 14.6 / 18.6 & 9.5 / 11.7 & 12.1 / 30.3 \\ 
VTransE (TDE)~\cite{sgg:tde}$\ _{\textit{CVPR'20}}$ & 48.5 / 53.1 & 24.6 / 28.0 & 36.6 / 40.6 & 25.7 / 28.5 & 12.9 / 14.8 & 19.3 / 21.7 & 18.7 / 22.6 & 8.6 / 10.5 & 13.7 / 16.6 \\ 
Motifs (TDE)~\cite{sgg:tde}$\ _{\textit{CVPR'20}}$  & 46.2 / 51.4 & 25.5 / 29.1 & 35.9 / 40.3 & 27.7 / 29.9 & 13.1 / 14.9 & 20.4 / 22.4 & 16.9 / 20.3 & 8.2 / 9.8 & 12.6 / 15.1 \\
VCTree (TDE)~\cite{sgg:tde}$\ _{\textit{CVPR'20}}$ & 47.2 / 51.6 & 25.4 / 28.7 & 36.3 / 40.2 & 25.4 / 27.9 & 12.2 / 14.0 & 18.8 / 21.0 & 19.4 / 23.2 & 9.3 / 11.1 & 14.4 / 17.2 \\ 
Motifs (CogTree)~\cite{sgg:cogtree}$\ _{\textit{IJCAI'21}}$  & 35.6 / 36.8 & 26.4 / 29.0 & 31.0 / 32.9 & 21.6 / 22.2 & 14.9 / 16.1 & 18.3 / 19.2 & 20.0 / 22.1 & 10.4 / 11.8 & 15.2 / 17.0 \\ 
VCTree (CogTree)~\cite{sgg:cogtree}$\ _{\textit{IJCAI'21}}$ & 44.0 / 45.4 & 27.6 / 29.7 & 35.8 / 37.6 & 30.9 / 31.7 & 18.8 / 19.9 & 24.9 / 25.8 & 18.2 / 20.4 & 10.4 / 12.1 & 14.3 / 16.3 \\
SG-Trans (CogTree)~\cite{sgg:cogtree}$\ _{\textit{IJCAI'21}}$  & 38.4 / 39.7 & 28.4 / 31.0 & 33.4 / 35.4 & 22.9 / 23.4 & 15.7 / 16.7 & 19.3 / 20.1 & 19.5 / 21.7 & 11.1 / 12.7 & 15.3 / 17.2 \\
Motifs (BA-SGG)~\cite{sgg:ba-sgg}$\ _{\textit{ICCV'21}}$ & 50.7 / 52.5 & 29.7 / 31.7 & 40.2 / 42.1 & 30.1 / 31.0 & 16.5 / 17.5 & 23.3 / 24.3 & 23.0 / 26.9 & 13.5 / 15.6 & 18.3 / 21.3 \\
VCTree (BA-SGG)~\cite{sgg:ba-sgg}$\ _{\textit{ICCV'21}}$ & 50.0 / 51.8 & 30.6 / 32.6 & 40.3 / 42.2 & \underline{34.0} / \underline{35.0} & 20.1 / 21.2 & \underline{27.1} / \underline{28.1} & 21.7 / 25.5 & 13.5 / 15.7 & 17.6 / 20.6 \\ 
Motifs (RTPB)~\cite{sgg:rtpb}$\ _{\textit{AAAI'22}}$ & 40.4 / 42.5 & 35.3 / 37.7 & 37.9 / 40.1 & 26.0 / 26.9 & 20.0 / 21.0 & 23.0 / 24.0 & 19.0 / 22.5 & 13.1 / 15.5 & 16.1 / 19.0 \\
VCTree (RTPB)~\cite{sgg:rtpb}$\ _{\textit{AAAI'22}}$ & 41.2 / 43.3 & 33.4 / 35.6 & 37.3 / 39.5 & 28.7 / 30.0 & 24.5 / 25.8 & 26.6 / 27.9 & 18.1 / 21.3 & 12.8 / 15.1 & 15.5 / 18.2 \\
DTrans (RTPB)~\cite{sgg:rtpb}$\ _{\textit{AAAI'22}}$ & 45.6 / 47.5 & 36.2 / 38.1 & 40.9 / 42.8 & 24.5 / 25.5 & 21.8 / 22.8 & 23.2 / 24.2 & 19.7 / 23.4 & 16.5 / 19.0 & 18.1 / 21.2 \\ 
Motifs (GCL)~\cite{sgg:gcl}$\ _{\textit{CVPR'22}}$ & 42.7 / 44.4 & 36.1 / 38.2 & 39.4 / 41.3 & 26.1 / 27.1 & 20.8 / 21.8 & 23.5 / 24.5 & 18.4 / 22.0 & 16.8 / 19.3 & 17.6 / 20.7 \\ 
VCTree (GCL)~\cite{sgg:gcl}$\ _{\textit{CVPR'22}}$ & 40.7 / 42.7 & 37.1 / 39.1 & 38.9 / 40.9 & 27.7 / 28.7 & 22.5 / 23.5 & 25.1 / 26.1 & 17.4 / 20.7 & 15.2 / 17.5 & 16.3 / 19.1 \\ 
IMP (PPDL)~\cite{sgg:ppdl} $\ _{\textit{CVPR'22}}$ & 39.5 / 39.7 & 24.8 / 25.3 & 32.2 / 32.5 & 25.8 / 26.7 & 14.2 / 15.9 & 20.0 / 21.3 & 18.5 / 19.4 & 9.8 / 10.4 & 14.2 / 14.9 \\ 
Motifs (PPDL)~\cite{sgg:ppdl}$\ _{\textit{CVPR'22}}$ & 47.2 / 47.6 & 32.2 / 33.3 & 39.7 / 40.5 & 28.4 / 29.3 & 17.5 / 18.2 & 23.0 / 23.8 & 21.2 / 23.9 & 11.4 / 13.5 & 16.3 / 18.7 \\ 
VCTree (PPDL)~\cite{sgg:ppdl}$\ _{\textit{CVPR'22}}$ & 47.6 / 48.0 & 33.3 / 33.8 & 40.5 / 40.9 & 32.1 / 33.0 & 21.8 / 22.4 & 27.0 / 27.7 &20.1 / 22.9 & 11.3 / 13.3 & 15.7 / 18.1  \\

\hline
\textbf{Transformer (DHL)} & 49.0 / 51.1 & \textbf{40.4 / 42.6} & 44.7 / 46.9 & 28.1 / 29.1 & \underline{24.2} / \underline{25.3} & 26.2 / 27.2 & 23.2 / \underline{27.3} & \textbf{18.2 / 21.0} & \underline{20.7} / \underline{24.2}\\
\textbf{Motifs (DHL)} & \underline{51.8} / \underline{53.8} & 39.1 / 41.7 & \underline{45.5} / \underline{47.8} & 27.4 / 31.1 & 23.1 / 24.2 & 25.3 / 27.7 & \textbf{24.7 / 28.8} & \underline{17.8} / \underline{20.7} & \textbf{21.3 / 24.8} \\ 
\textbf{VCTree (DHL)} & \textbf{52.3 / 54.2} & \underline{40.0} / \underline{42.2} & \textbf{46.2 / 48.2} & \textbf{36.6 / 37.8} & \textbf{26.9 / 28.2} & \textbf{31.8 / 33.0} & \underline{23.3} / 27.1 & 17.4 / 20.0 & 20.4 / 23.6 \\
\hline
\end{tabular}
}
\label{tab:R@k_mR@k} 
\end{table*}

\subsection{Implementation Details}
\noindent
\textbf{Object Detector.}
Following the previous works, we adopt the Faster R-CNN~\cite{fasterrcnn} with ResNeXt-101-FPN~\cite{fpn,resxnet,resnet} as the backbone pre-trained by~\cite{sgg:sggbenchmark} to detect objects in the image. The parameters of the network are frozen during the scene graph generation training.

\noindent
\textbf{Scene Graph Generation.}
Our proposed method is model-agnostic so that it can be integrated with most existing scene graph models. Therefore, we select three strong baselines in Model Zoo~\cite{sgg:sggbenchmark}: \textbf{Transformer}~\cite{sgg:sggbenchmark,transformer}, \textbf{Motifs}~\cite{sgg:motifs}, and \textbf{VCTree}~\cite{sgg:vctree} to evaluate the effectiveness of our method. The hyper-parameters of them are identical to the setting in Model Zoo~\cite{sgg:sggbenchmark}. 
In the Curriculum Re-weighting Mechanism, the $\beta_1$ is set to 0.2, and the $M$ is set to 10,000. 
The $D$ in Semantic Context Module is set to 512. 
For the Branch Curriculum Schedule, the $\beta_2$ is set to 0.1, and the $K_1$ and $K_2$ are set to 10,000 and 20,000.
For the Knowledge Distillation, the temperature $\tau$ is set to 2.
The balancing factor $\mu$ is set to 0.05. 
All models are trained by an SGD optimizer with 40K iterations. The batch size and learning rate are set to 12 and $12 \times 10^{-3}$. 

\noindent
\textbf{Sentence-to-Graph Retrieval.} 
We follow the same formulation of Sentence-to-Graph Retrieval in~\cite{sgg:tde}. This task is formulated as a matching problem between image caption and scene graph. The captions of the images are converted to text graphs by method~\cite{sg-parser}. and we use the Bilinear~\cite{bilinear} attention network to map the text graphs and scene graphs into an embedding space. The batch size and the learning rate are set to 12 and $12 \times 10^{-2}$. 

\noindent
\textbf{Image Captioning.} 
Our implementation is based on the code base~\cite{cap:codebase}. We use the Transformer Captioning model as the baseline for the image captioning task. We use the SGG model trained on VG dataset to extract scene graphs in the MS-COCO dataset for the image captioning task. 
We selected the 128 highest scoring triplets in an image. For each triplet, we use a pre-trained word embedding model (GloVe) to map the subject, predicate, and object to three 300-dimensional vectors, respectively, then concatenate them to a 900-dimensional vector as the representation of this triplet.
The triplet representations in an image are injected into the Captioning model as additional auxiliary information.
The model trained 20 epochs with the batch size of 5 and the learning rate $5 \times 10^{-4}$. In the inference stage, the beam size is 5. 

\subsection{Experimental Results}
\noindent
\textbf{Scene Graph Generation.} 
We evaluate our DHL by incorporating it into three typical baseline models, namely Motifs~\cite{sgg:motifs}, VCTree~\cite{sgg:vctree}, and Transformer~\cite{sgg:sggbenchmark,transformer}.
For a fair comparison with existing state-of-the-art model-agnostic de-biasing methods, including TDE~\cite{sgg:tde}, EBM~\cite{sgg:ebm}, CogTree~\cite{sgg:cogtree}, BA-SGG~\cite{sgg:ba-sgg}, RTPB~\cite{sgg:rtpb}, FGPL~\cite{sgg:fgpl}, GCL~\cite{sgg:gcl}, and PPDL~\cite{sgg:ppdl}, we also incorporate them into these three baselines, respectively.
The comparison results are shown in Table~\ref{tab:mR@k}.
From Table~\ref{tab:mR@k}, we have the following observations:
1) Compared with the three strong baselines, our DHL consistently improves the model performance by more than 100\% over three tasks, \eg we improve the vanilla VCTree from 17.4 to 42.2 on mR@100 PredCls, from 11.5 to 28.2 on mR@100 SGCls, and from 7.9 to 20.0 on mR@100 SGDet,
2) Compared with other model-agnostic de-biasing strategies, DHL achieves the best performance, \eg Motifs (DHL), VCTree (DHL), and Transformer (DHL) outperform the latest state-of-the-art methods Motifs (FGPL), VCTree (FGPL), and Transformer (FGPL) with consistent improvements as 8.6, 2.5, and 7.0 on mR@20 PredCls.
Experimental results demonstrate that our proposed DHL significantly improves the ability of the vanilla SGG model to predict more informative tail predicates. 

%
%
%

Moreover, to comprehensively compare the performance with existing de-biasing methods, we report the R@50/100, mR@50/100, and M@50/100 on three tasks in Table~\ref{tab:R@k_mR@k}. 
From Table~\ref{tab:R@k_mR@k}, we can observe that many de-biasing methods achieve relatively competitive performance in mR@K metrics, but their R@K metrics are weak, \eg Motifs (Cogtree), Motifs (RTPB) and Motifs (GCL) obtain 29.0, 37.7, and 38.2 in mR@100, but only have 36.8, 42.5, and 44.4 in R@100 on the PredCls task.
Compared with these de-biasing methods, our methods obtain excellent performance in mR@K metrics and maintain relatively satisfactory performance in R@K metrics, \eg VCTree (DHL) achieves 42.2 in mR@100 and 54.2 in R@100.
It is worth mentioning that all our methods outperform all these existing de-biasing methods in R@K and mR@K on the challenging SGDet task. 
The results show that our methods achieve a satisfactory performance balance between head predicates and tail predicates.

\begin{figure*}[!t]
  \center
  \includegraphics[width=1\linewidth]{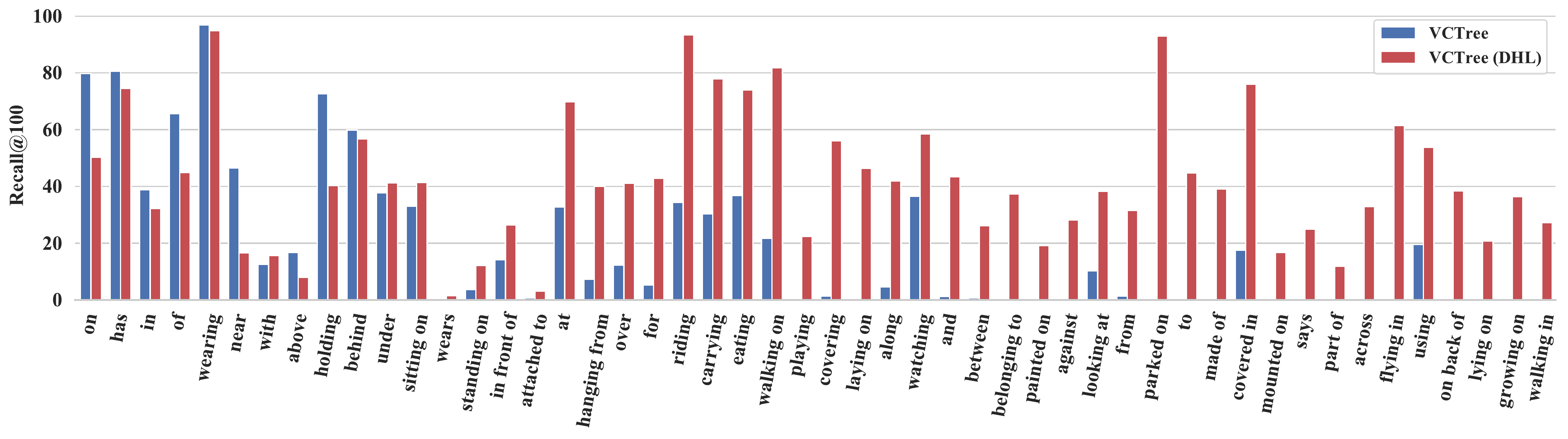}
\caption{R@100 of all the predicate class of vanilla VCTree and VCTree (DHL) on the PredCls. Predicates are sorted in decreasing order of the sample frequency.
}
\label{fig:dist}
\end{figure*}

To study the performance of predicates with different popularity, we sort the 50 predicates by their sample frequencies and divide them into three groups, Many (17), Medium (17), and Few (16). Table ~\ref{tab:three_group} shows the Group Mean Recall@100 on these groups under the PredCls task. 
Generally, after being integrated with DHL, we observe significant improvements in the Medium and Few groups compared with baselines, \eg boosting VCTree from 8.0 to 47.7 in the Medium group and from 3.1 to 40.4 in the Few group. 
At the same time, we maintain the performance of head predicates, which drops slightly (only 1.9 in Many Group with VCTree(DHL)).
Moreover, we show the Recall@100 of each predicate for vanilla VCTree and VCTree (DHL) on the PredCls task in Fig.~\ref{fig:dist}. 
From Fig.~\ref{fig:dist}, we observe that the performance degradation of head predicates mainly focuses on ``on'', ``of'', ``near'', and ``holding''.
The reason is that VCTree (DHL) reasonably classifies some coarse-grained head predicates as fine-grained tail predicates (\eg refining ``on'' into ``walking on'', ``near'' into ``on back of'').
As for the tail predicates, we obtain huge enhancements, some even improving from 0 Recall@100 (\eg ``parked on'', ``flying in'', and ``growing on''). The results further prove the superiority of our method in predicting informative tail predicates.

In addition to the VG dataset, we verify our method on a more challenging dataset, GQA. 
As shown in Table~\ref{tab:GQA}, our methods achieve almost 300\% performance improvements on three tasks compared to the baseline models.
For instance, our DHL improves Motifs from 4.6 to 21.9 and Transformer from 5.0 to 20.1 in mR@100 metric on the PredCls task. 
The results demonstrate that our method has good generalization and can handle complex scenes.

\begin{table}[!t]
\caption{Performance comparison in Group Mean Recall@100 on the PredCls task.}
    \centering
    \begin{tabular}{c|ccc}
     \hline
     \multirow{2}{*}{Methods} & \multicolumn{3}{|c}{PredCls} \\ 
     \cline{2-4}
     & Many (17) & Medium (17) & Few (16)\\ 
     \hline
     Motifs & \textbf{40.0} & 9.8 & 2.5  \\ 
     \textbf{Motifs (DHL)} & 36.9 & \textbf{50.2} & \textbf{37.7} \\ 
     \hline
     VCTree & \textbf{40.1} & 8.0 & 3.1  \\
     \textbf{VCTree (DHL)} & 38.2 & \textbf{47.7} & \textbf{40.4} \\ 
     \hline
     Transformer & \textbf{38.3} & 9.4 & 3.1 \\
     \textbf{Transformer (DHL)} & 35.5 & \textbf{51.1} & \textbf{41.0} \\ 
     \hline
    \end{tabular}
\label{tab:three_group}
\end{table}

\begin{table}[!t] \centering 
\caption{Performance comparison of different methods in mR@50/100 on the PredCls task for the GQA dataset. Trans means the Transformer SGG model.}
\resizebox{\linewidth}{!}{
    \begin{tabular}{c|ccc}
    \hline
    \multirow{2}{*}{Models} & PredCls & SGCls & SGDet   \\
    \cline{2-4}
    & mR@50 / 100 & mR@50 / 100 & mR@50 / 100  \\
    \hline
    Motifs & 4.1 / 4.6 & 1.8 / 1.9 & 1.8 / 2.1 \\ 
    \textbf{Motifs (DHL)}  & \textbf{20.4} / \textbf{21.9} & \textbf{8.4} / \textbf{9.1} & \textbf{6.6} / \textbf{8.1} \\
    \hline
    Trans & 4.6 / 5.0 & 2.2 / 2.3 & 1.7 / 2.0 \\ 
    \textbf{Trans (DHL)} & \textbf{18.2} / \textbf{20.1} & \textbf{8.7} / \textbf{9.3} & \textbf{7.8} / \textbf{8.8} \\ 
    \hline
    \end{tabular}
}
\label{tab:GQA} 
\end{table}

\begin{table}[!t]
    \caption{Performance comparison of Sentence-to-Graph Retrieval. Trans means the Transformer SGG model.}
    \centering
    \resizebox{\linewidth}{!}{
    \begin{tabular}{c|ccc|ccc}
    \hline
    \multirow{2}{*}{Model} & \multicolumn{3}{c|}{Gallery 1000} & \multicolumn{3}{c}{Gallery 5000} \\
    \cline{2-7}
    & R@20 & R@50 & R@100 & R@20 & R@50 & R@100 \\ 
    \hline
     Motifs & 15.7 & 29.8 & 45.9 & 4.1 & 8.5 & 15.6 \\ 
    \textbf{Motifs (DHL)} & \textbf{24.9} & \textbf{46.0} & \textbf{61.3} & \textbf{7.8} & \textbf{17.1} & \textbf{28.7} \\ 
    \hline
     Trans  & 15.0 & 28.3 & 44.8 &  3.7 & 8.0 & 14.4 \\
    \textbf{Trans(DHL)} & \textbf{22.5} & \textbf{44.2} & \textbf{62.6} & \textbf{7.1} & \textbf{15.5} & \textbf{26.3} \\
    
    \hline
    \end{tabular}
    }
    \label{tab:S2GR}
\end{table}

\begin{table}[!t]
    \caption{Performance comparison of Image Captioning. Trans means the Transformer SGG model.}
    \centering
    \resizebox{\linewidth}{!}{
    \begin{tabular}{c|cccc}
    \hline
     Model & Bleu-4 & Meteor & Cider & Spice \\ 
     \hline
     Baseline~\cite{cap:codebase} & 35.3 & 27.6 & 111.8 & 20.6 \\
     Baseline+Trans & 35.3 & 27.6 & 111.9 & 20.7 \\ 
     \textbf{Baseline+Trans (DHL)} & \textbf{35.3} & \textbf{27.8} & \textbf{112.9} & \textbf{20.7} \\ 
     \hline
    \end{tabular}
    }
    \label{tab:cap}
\end{table}

\noindent
\textbf{Sentence-to-Graph Retrieval.}
Table~\ref{tab:S2GR} shows the results of Sentence-to-Graph Retrieval. We use Motifs~\cite{sgg:motifs} and Transformer~\cite{transformer, sgg:sggbenchmark} as baseline models. After equipping with our DHL, we observe significant performance improvements. Such as, we improve Motifs from 45.9 to 61.3 in R@100 on Gallery 1000 and from 15.6 to 28.7 in R@100 on Gallery 5000.
The improvements can be attributed to the excellent performance of our method in generating unbiased scene graphs. 
Primarily, our method ensures the semantic consistency between the generated scene graph and the ground truth scene graph. This matches the original intent of the Sentence-to-Graph Retrieval task.

\noindent
\textbf{Image Captioning.}
We use the Transformer~\cite{sgg:sggbenchmark,transformer} trained on VG~\cite{data:vg,sgg:imp} dataset to generate scene graphs for the image captioning task. 
As shown in Table~\ref{tab:cap}, the baseline model~\cite{cap:codebase} with the scene graphs generated by the vanilla Transformer has few improvements. 
However, the baseline model with the scene graphs generated by Transformer (DHL) obtains the improvement on Cider (112.9 \vs 111.8). 
This shows the excellent ability of our method to generate informative scene graphs and verifies that the scene graphs can indeed improve the downstream tasks.

\subsection{Ablation Study}
In this subsection, extensive experiments are conducted to investigate each component’s contribution and possible variants in our proposed DHL. 
In particular, we use the Transformer~\cite{sgg:sggbenchmark, transformer} model as the baseline and only perform the task of PredCls for fast validation.

\noindent
\textbf{Effectiveness of each component.} 
To validate the effectiveness of each component of the DHL, we conduct the related ablation experiments based on the below settings:

\begin{enumerate}
    \item[(1)] 
    \textit{CLB.} 
    This model only performs the Coarse-grained Learning Branch (CLB), which is our baseline model, \ie vanilla Transformer~\cite{sgg:sggbenchmark,transformer} SGG model.
    
    \item[(2)] 
    \textit{FLB.} This model only performs the Fine-grained Learning Branch (FLB). Based on this, FLB (w/o CRM) and FLB (w/o SCM) mean that we remove the Curriculum Re-weighting Mechanism (CRM) (\ie using the standard cross-entropy loss to optimize this branch) and Semantic Context Module (SCM) from FLB.
    
    \item[(3)] 
    \textit{CLB+FLB.} This model is trained by combining the CLB and FLB.
    
    \item[(4)] 
    \textit{CLB+FLB+KD.} This model is our final model. Here, KD denotes the knowledge distillation. We distill the expert knowledge of CLB on head predicates to FLB for reliable head predicate classification.
\end{enumerate}

The experimental results are summarized in Table~\ref{tab:abl_each_components}.
From Table~\ref{tab:abl_each_components}, we obtain several observations as follows: 
\begin{enumerate}
    \item[(1)]
    Among all the above settings, CLB performs the worst in mR@K metrics but obtains excellent performance in R@K. Thus, it can provide expertise and robust features of head predicates for FLB. 
    
    \item[(2)]
    Compared with CLB, FLB achieves satisfactory performance in mR@K metrics (42.3 \vs 17.2 in mR@100), and exceeds all the comparing methods shown in Table ~\ref{tab:mR@k}. 
    This shows the superiority of our FLB in identifying tail predicates. 
    \item[(3)]
    CLB+FLB further improves the performance of mR@K (43.9 \vs 42.3 in mR@100) without sacrificing the performance of R@K (43.6 \vs 43.4 in R@100). 
    This confirms that robust head predicates features are indeed conducive to tail predicates learning.
    \item[(4)]
    CLB+FLB+KD achieves huge improvements in R@K (51.1 \vs 43.6 in R@100). This illustrates that expertise in head predicates of CLB can alleviate the ambiguous classifications of FLB on head predicates, which reduces the risk of over-fitting the tail predicates and makes the model more reliable. Besides, this model achieves the best results in M@K, illustrating the superiority of its comprehensive performance.
    \item[(5)]
    Compared with FLB (w/o-CRM), FLB achieves more than 100\% improvements in mR@K metrics (42.3 \vs 19.3 in mR@100), which verifies that CRM can fully explore and utilize the general patterns provided by head predicates to learn tail predicates efficiently.
    Compared with FLB (w/o-SCM), FLB obtains comprehensive improvements in R@K and mR@K, demonstrating that SCM takes advantage of the context to correct the out-of-context predictions into appropriate predicates.
    The refined predicates may contain both head predicates and tail ones.
\end{enumerate}

\begin{table}[!t]
\caption{Ablation study of the effectiveness of our proposed components. \textit{CLB}, \textit{FLB}, \textit{KD} denote coarse-grained learning branch, fine-grained learning branch and knowledge distillation, respectively. ``w/o'' means to remove a sub-module from the basic model. The baseline is vanilla Transformer~\cite{transformer,sgg:sggbenchmark} SGG model.}
    \centering
    \resizebox{\linewidth}{!}{
    \begin{tabular}{c|c|c|c}
         \hline
        \multirow{2}{*}{Ablation Models}  & \multicolumn{3}{c}{Predcls} \\
         \cline{2-4}
         & M@50/100 & mR@50/100 & R@50/100 \\ 
         \hline
         CLB (Transformer) & 40.6 / 42.1 & 15.9 / 17.2 & 65.2 / \textbf{66.9} \\
         \hline
         FLB (w/o CRM) & 41.5 / 43.1 & 17.6 / 19.3 & \textbf{65.4} / \textbf{66.9} \\ 
         FLB (w/o SCM) & 39.4 / 41.3 & 38.7 / 40.4 & 40.1 / 42.2  \\ 
         FLB & 41.0 / 42.9 & 40.2 / 42.3 & 41.8 / 43.4 \\ 
         \hline
         CLB+FLB  & 41.5 / 43.8 & \textbf{41.4} / \textbf{43.9} & 41.5 / 43.6 \\ 
         FLB+CLB+KD &\textbf{44.7} / \textbf{46.9} & 40.4 / 42.6 & 49.0 / 51.1 \\ 
         \hline
    \end{tabular}
    }
    \label{tab:abl_each_components}
\end{table}

\noindent
\textbf{Variants to Schedule Function.}
In the Predicate Curriculum Schedule and Branch Curriculum Schedule, \textbf{Schedule Functions} $\varphi_1(\cdot)$ and $\varphi_2(\cdot)$ are used to adjust the learning focus in the training process. They have a simple formula as follows:
\begin{equation}
    \varphi(t) = 1 - \frac{t}{T},
\end{equation}
where $T$ is a constant and $t$ is a variable.
In order to explore the impact of different decreasing functions on the learning process, we test the following functions:
\begin{enumerate}
    \item[(1)] Exponential function, indicates the speed of transfer from fast to slow, defined as:
    \begin{equation}
        \varphi(t) = \nu^{\frac{t}{T}}, \text{($0 < \nu < 1$)}.
    \end{equation}
    \item[(2)] Parabolic function, indicates the speed of transfer from slow to fast, defined as:
    \begin{equation}
        \varphi(t) = 1 - (\frac{t}{T})^2.
    \end{equation}
\end{enumerate}
The experimental results of different schedule functions in Predicate Curriculum Schedule are shown in Table~\ref{tab:abl_func}. From the results, we have several observations as follows: 
1) Parabolic function is superior to linear function and exponential function in R@K metrics but inferior to them in mR@K metrics. 
The possible reason is that the transfer speed of learning focus of parabolic function is from slow to fast, which leads to more focus on learning head predicates and less on learning the tail predicates.
2) Exponential function has the worst head predicates performance. This is because exponential function quickly shifts the learning focus from the head predicates to the tail predicates, resulting in insufficient learning of the head predicates.
3) The linear function has a constant transfer speed, which shifts the learning focus from the head predicates to the tail predicates smoothly and achieves the best results in mR@K and M@K. 
Therefore, we choose the linear function as the schedule function in Predicate Curriculum Schedule. 
Similarly, we also use the linear function in Branch Curriculum Schedule. 

\begin{table}[!t]
    \caption{Ablation study of different schedule functions in Predicate Curriculum Schedule. Exp means Exponential function, Par means Parabolic function, and Lin means Linear function. The results are obtained with Transformer(DHL) model.}
    \centering
     \resizebox{\linewidth}{!}{
    \begin{tabular}{c|cc|cc|cc}
    \hline
    \multirow{2}{*}{Func} & \multicolumn{6}{c}{PredCls} \\
    \cline{2-7}
    & M@50 & M@100 & mR@50 & mR@100 & R@50 & R@100 \\ 
    \hline
    Exp  & 44.1 & 45.6 & 40.0 & 41.7 & 48.1 & 49.4 \\
    Par  & 44.5 & 46.3 & 39.5 & 41.3 & \textbf{49.4} & \textbf{51.2} \\  
    Lin  & \textbf{44.7} & \textbf{46.9} & \textbf{40.4} & \textbf{42.6} & 49.0 & 51.1 \\
    \hline
    \end{tabular}
    }
    \label{tab:abl_func}
\end{table}


\begin{table}[!t]
    \caption{Ablation study of different constructions for the semantic representation of scene graph in SCM. The results are obtained with Transformer (DHL) model.}
    \centering
     \resizebox{\linewidth}{!}{
    \begin{tabular}{c|cc|cc|cc}
    \hline
    \multirow{2}{*}{Method} & \multicolumn{6}{c}{PredCls} \\
    \cline{2-7}
    & M@50 & M@100 & mR@50 & mR@100 & R@50 & R@100 \\ 
    \hline
    Mean & 44.2 & 45.5 & 40.0 & 41.5 & 48.3 & 49.5 \\
    Global  & \textbf{44.7} & \textbf{46.9} & \textbf{40.4} & \textbf{42.6} & \textbf{49.0} & \textbf{51.1} \\
    \hline
    \end{tabular}
    }
    \label{tab:abl_scm}
\end{table}



\begin{table}[!]
\caption{Hyper-parameter analysis of the number of head samples $M$. The results are obtained with Transformer (DHL) model.}
    \centering
    \resizebox{\linewidth}{!}{
    \begin{tabular}{c|cc|cc|cc}
    \hline
    \multirow{2}{*}{$M$} & \multicolumn{6}{c}{PredCls} \\
    \cline{2-7}
    & M@50 & M@100 & mR@50 & mR@100 & R@50 & R@100 \\
    \hline
    5,000  & 44.5 & 46.6 & 39.5 & 41.4 & \textbf{49.4} & \textbf{51.8} \\ 
    10,000 & \textbf{44.7} & \textbf{46.9} & \textbf{40.4} & \textbf{42.6} & 49.0 & 51.1 \\
    40,000 & 44.1 & 46.4 & 40.3 & 42.4 & 47.9 & 50.1 \\
    \hline
    \end{tabular}
    }
\label{tab:analy_m}
\end{table}



\begin{table}[!]
\caption{Parameter analysis of distillation temperature $\tau$. The results are obtained with Transformer (DHL) model.}
    \centering
    \begin{tabular}{c|cc|cc|cc}
    \hline
    \multirow{2}{*}{$\tau$} & \multicolumn{6}{c}{PredCls} \\
    \cline{2-7}
    & M@50 & M@100 & mR@50 & mR@100 & R@50 & R@100 \\
    \hline
    1 & 44.6 & 46.4 & 38.9 & 40.8 & \textbf{50.2} & \textbf{52.0} \\
    2 & \textbf{44.7} & \textbf{46.9} & 40.4 & 42.6 & 49.0 & 51.1 \\
    3 & 43.2 & 45.4 & 41.2 & 43.5 & 45.1 & 47.2  \\
    4 & 41.9 & 44.2 & 41.2 & 43.7 & 42.6 & 44.7 \\
    5 & 41.6 & 44.0 & \textbf{41.9} & \textbf{44.5} & 41.3 & 43.4 \\ 
    \hline
    \end{tabular}
    \label{tab:ana_tau}
\end{table}

\begin{figure*}[t]
 \centering
 \includegraphics[width=1.0\linewidth]{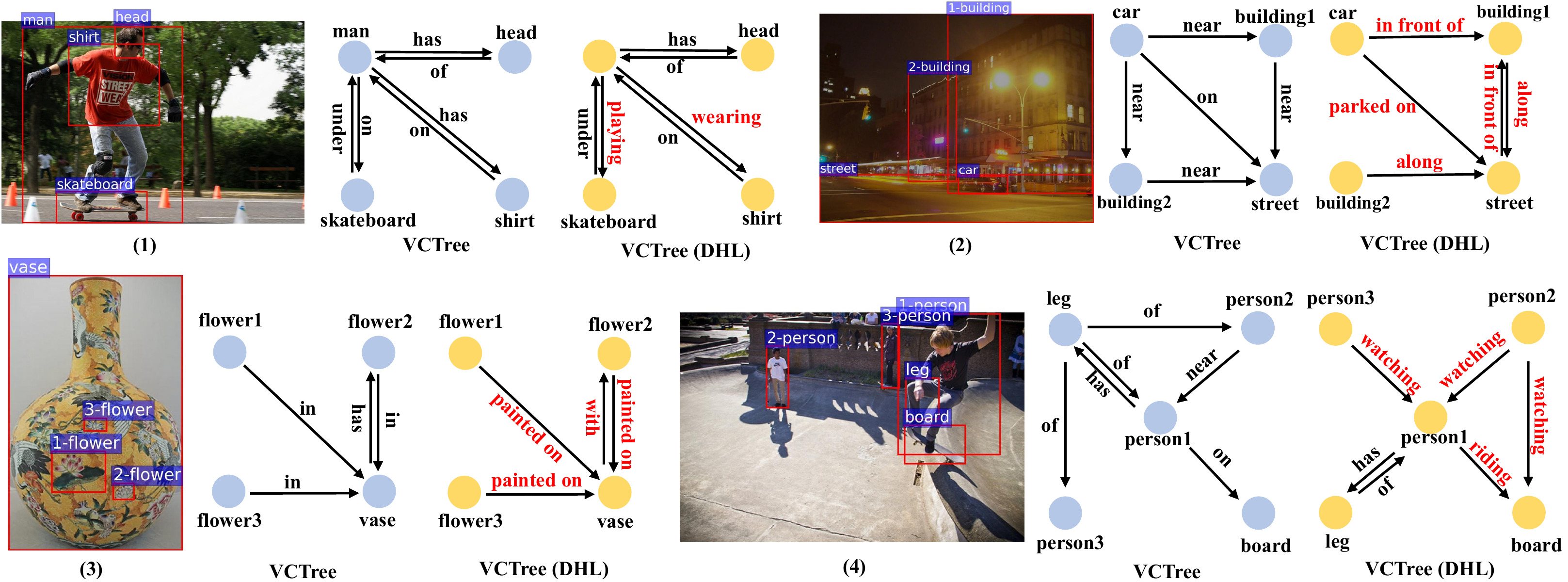} 
 \caption{Visualization Results of VCTree in blue and VCTree (DHL) in yellow on the PredCls task. The scene graph generated with the VCTree (DHL) is more informative than the one generated with the vanilla VCTree. }
 \label{fig:ksh}
\end{figure*}
\noindent
\textbf{Variants to Semantic Context Module.} 
In this module, an extra global representation $s_{global}$ (\textbf{Global}) mentioned in Sec.~\ref{sec:SCM} is utilized as the semantic representation of the scene graph. 
In order to demonstrate the effectiveness of $s_{global}$, we first remove $s_{global}$ in $S^r$, and then take the average of $\{\widetilde{s_1^r},\widetilde{s_2^r},...,\widetilde{s_N^r}\}$ (\textbf{Mean}) as $\widetilde{s}_{global}$ in Eq.~(\ref{eq:sc_loss}). The ground truth performs the same operations.
Shown in Table~\ref{tab:abl_scm}, \textbf{Global} exceeds \textbf{Mean} in all metrics, which shows the superiority of the global representation $s_{global}$. 
For most SGG methods, any two object proposals in an image will predict a relation predicate and form a relationship triplet. All relationship triplets in an image are sent to SCM to construct the context semantics representations, but most of them are noises. 
If we use \textbf{Mean} as the semantic representation of the scene graph, these noises are added. 
However, the extra \textbf{Global} is flexible, which can aggregate the semantics of appropriate relationship triplets to form the semantic representation of the scene graph to avoid the mixing of noise. 
Thus, it is appropriate to use the \textbf{Global} as the semantic representation of the scene graph.

\subsection{Hyper-parameter Analysis}
In this subsection, we analyze several essential hyper-parameters in our method. 
How to divide the head predicates and tail predicates is essential. In Sec.~\ref{sec:CRM}, we define the head predicates with the number of samples exceeding $M$. 
In Table~\ref{tab:analy_m}, we evaluate three different $M$, and the results are analyzed as follows.
When $M=5,000$, there are 22 head predicates. Thus, more knowledge is distilled from CLB to FLB, improving the R@K metrics that represent the head predicates' performance. However, some predicates that CLB learns poorly are also distilled to FLB, which disturbs the decisions of FLB, resulting in the decrease of mR@K metrics.
When $M=40,000$, knowledge of only 10 predicates is distilled from CLB to FLB, resulting in a waste of knowledge and a drop in R@K metrics. 
When $M=10,000$, there are 16 head predicates, and the model achieves relatively satisfactory results in R@K and the best performance in mR@K and M@K.

Moreover, the distillation temperature $\tau$ in Eq.~(\ref{eq:kd_t}) is also a vital hyper-parameter.
As~\cite{kd1} mentioned, when the temperature $\tau$ increases, the soft probability distribution $p^T$ and $q^T$ in Eq.~(\ref{eq:loss_kd}) are more balanced. When $\tau \rightarrow \infty$, $q^T$ and $p^T$ become the uniform distribution.
As shown in Table~\ref{tab:ana_tau}, with the increase of temperature $\tau$, the R@K metrics gradually decrease, while the mR@K metrics increase.
Therefore, a high temperature weakens the effect of knowledge distillation because the soft probability distribution only contains little information. 
When $\tau = 2$, the model achieves the best performance in M@K.  
Based on these analyses, we use $M=10,000$ and $\tau=2$ in our DHL.

\subsection{Visualization Results}
We visualize scene graphs generated by the vanilla VCTree and VCTree (DHL) in Fig.~\ref{fig:ksh}. 
It is evident that VCTree (DHL) generates more fine-grained predicates than the vanilla VCTree, 
\eg (man, playing, skateboard) \vs (man, on, skateboard) in the first example, 
(car, parked on, street) \vs (car, on, street) and (car, in front of building1) \vs (car, near, building1) in the second example, 
(flower1, painted on, vase) \vs (flower1, in, vase) in the third example. 
In the fourth example, the vanilla VCTree predicts some unreasonable relationships, \eg (leg, of, person2), (leg, of, person3), and (leg, of, person1). However, VCTree (DHL) corrects these unreasonable predictions. 
We think the possible reason is that the SCM uses the context information, (person1, has, leg), to infer that the ``leg'' can only belong to the ``person1''. 
Furthermore, VCTree (DHL) generates some meaningful interactions, \eg (person2, watching, person1), (person2, watching, board) and (person1, riding, board). 
These results demonstrate the effectiveness of our method in generating unbiased scene graphs.

\section{CONCLUSION}
\label{sec:CON}
In this work, we propose a novel Dual-branch Hybrid Learning network (DHL) for unbiased scene graph generation.
In particular, our DHL is composed of a Coarse-grained Learning Branch (CLB) for learning expertise and robust features of head predicates and a Fine-grained Learning Branch (FLB) for predicting informative tail predicates.
To make the two branches work well together, we design a Branch Curriculum Schedule (BCS) to regulate the model to learn CLB first and then gradually pay attention to the learning of FLB.
Moreover, we introduce a Curriculum Re-weighting Mechanism (CRM) to optimize the FLB that first explores the general patterns from the head (easy) predicates and then gradually focuses on learning the tail (hard) predicates.
Besides, we devise a Semantic Context Module (SCM) to correct the out-of-context predictions, and the SCM can also ensure the semantic consistency between generated scene graph and ground truth at the graph level.
Extensive experiments on VG dataset and more complex GQA dataset verify the superiority and generalization of our method in predicting informative tail predicates.
The experimental results on two downstream tasks demonstrate our method's practicability of generated scene graphs.

\bibliographystyle{IEEEtran}
\bibliography{ref}

\end{document}